\documentclass[runningheads]{llncs}

\usepackage{eccv}

\usepackage{eccvabbrv}

\usepackage{booktabs}
\usepackage{color}
\usepackage{diagbox}
\usepackage{microtype}
\usepackage{makecell}

\usepackage{colortbl}
\usepackage{url}
\usepackage{}

\makeatletter

\newcommand{\Rmnum}[1]{\textcolor{red}{\expandafter\@slowromancap\romannumeral #1@}}
\makeatother

\usepackage{graphicx}
\usepackage{booktabs}

\usepackage[accsupp]{axessibility}  

\usepackage{hyperref}
\usepackage[capitalize]{cleveref}
\crefname{section}{Sec.}{Secs.}
\Crefname{section}{Section}{Sections}
\Crefname{table}{Table}{Tables}
\crefname{table}{Tab.}{Tabs.}

\usepackage{orcidlink}

\usepackage{nth}
\usepackage{makecell}
\usepackage{xcolor}
\usepackage{multirow}
\usepackage{enumerate}

\usepackage[capitalize]{cleveref}
\crefname{section}{Sec.}{Secs.}
\Crefname{section}{Section}{Sections}
\Crefname{table}{Table}{Tables}
\crefname{table}{Tab.}{Tabs.}

\def\pca{HPPM}
\def\pcaf{Human Part Parametric Models}

\def\dsf{Partially Visible}
\def\methodf{Divide and Fuse}
\def\method{D\&F}
\def\hmf{\textit{Partially Visible Human3.6M}}
\def\hm{\textit{PV-Human3.6M}}
\def\pwf{\textit{Partially Visible 3DPW}}
\def\pw{\textit{PV-3DPW}}

\newcommand{\supl}[1]{Supplementary Material Sec. \textcolor{red}{#1}}

\begin{document}

\title{Divide and Fuse: Body Part Mesh Recovery 
from \\ Partially Visible Human Images} 

\titlerunning{Divide and Fuse}

\author{Tianyu Luan\inst{2,1}\thanks{This work was carried out during the internship of Tianyu Luan at United Imaging Intelligence, Boston MA.}\orcidlink{0000-0001-7333-1052} \and
Zhongpai Gao\inst{1}\orcidlink{0000-0003-4344-4501} \and
Luyuan Xie\inst{3} \and
Abhishek Sharma\inst{1} \and
Hao Ding\inst{4} \and
Benjamin Planche \inst{1} \orcidlink{0000-0002-6110-6437} \and 
Meng Zheng \inst{1} \and 
Ange Lou \inst{5} \and 
Terrence Chen \inst{1} \and 
Junsong Yuan \inst{2} \orcidlink{0000-0002-7901-8793} \and 
Ziyan Wu \inst{1} \orcidlink{0000-0002-9774-7770}
}

\authorrunning{Luan. et al.}

\institute{United Imaging Intelligence, Boston MA, USA \and 
State University of New York at Buffalo, Buffalo NY 14260, USA \and
Peking University, Beijing, China \and
Johns Hopkins University, Baltimore MD 21218, USA \and
Vanderbilt University, Nashville TN 37235, USA \\
\email{\{tianyulu,jsyuan\}@buffalo.edu, \{first.last\}@uii-ai.com,} \\
\email{2201110745@stu.pku.edu.cn, hding15@jhu.edu, ange.lou@vanderbilt.edu.}
}

\maketitle

\begin{abstract}
We introduce a novel bottom-up approach for human body mesh reconstruction, specifically designed to address the challenges posed by partial visibility and occlusion in input images. Traditional top-down methods, relying on whole-body parametric models like SMPL, falter when only a small part of the human is visible, as they require visibility of most of the human body for accurate mesh reconstruction. To overcome this limitation, our method employs a ``\methodf{} (\method{})'' strategy, reconstructing human body parts independently before fusing them, thereby ensuring robustness against occlusions. We design \pcaf{} (\pca{}) that independently reconstruct the mesh from a few shape and global-location parameters, without inter-part dependency. A specially designed fusion module then seamlessly integrates the reconstructed parts, even when only a few are visible. We harness a large volume of ground-truth SMPL data to train our parametric mesh models. To facilitate the training and evaluation of our method, we have established benchmark datasets featuring images of partially visible humans with \pca{} annotations. Our experiments, conducted on these benchmark datasets, demonstrate the effectiveness of our \method{} method, particularly in scenarios with substantial invisibility, where traditional approaches struggle to maintain reconstruction quality.
\end{abstract}

\begin{figure}
    \centering
    \includegraphics[width=1.0\linewidth]{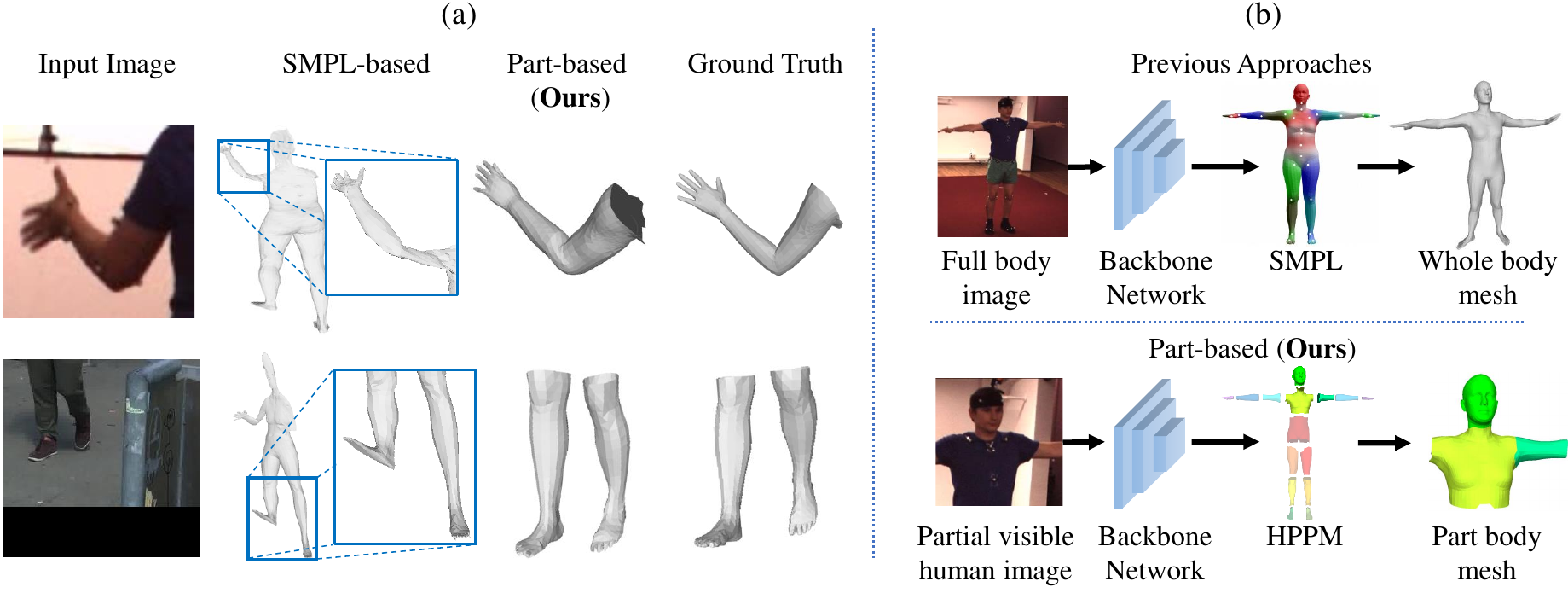}
    \caption{Traditional top-down method vs. \methodf{}. (a) When the input image only shows a few body parts (\nth{1} column), top-down SMPL-based methods may easily fail (\nth{2} column) due to the lack of whole-body information. Our part-based \method{} method is designed for partially visible human reconstruction (see results in the \nth{3} column). (b) Primary framework of SMPL-based prior art versus our proposed model.
    }
    \label{fig:teaser}
\end{figure}

\section{Introduction}

Human body mesh recovery has applications in various fields including augmented and virtual reality (AR/VR), film production, human-computer interaction (HCI), and sports. In specific applications such as movies, video games, or medical in-bore cameras, there are instances when major portions of the human body are outside the camera's field of view, leaving only small parts of the body visible. This scenario poses a significant challenge in accurately reconstructing the human mesh.

Previous methods, such as \cite{ma20233d,motionbert2022,nam2023cyclic, yang2023sefd, li2022cliff, li2023niki, pymaf2021, pymafx2023}, are effective when the majority of the human body is visible, but their performance significantly decreases when the human body is substantially invisible. Previous human mesh reconstruction methods mainly follow top-down designs, utilizing whole-body parametric models such as SMPL~\cite{SMPL-X:2019} or STAR~\cite{STAR:2020}. These models extract global features from input images and transform them into parameters to reconstruct the human mesh, and are successful when the entire body is nearly fully visible. However, when the input human body is largely incomplete, the human body may not be well recognized by the network. Moreover, when using top-down parametric models, the parameters are entangled among different body parts, making the reconstruction of visible sections dependent on the information from the entire body.  The absence of invisible parts will compromise this reliance and entanglement, thereby influencing the visible parts' reconstruction quality. Both the misrecognition and unnecessary entanglement issue would cause a drop in the body mesh accuracy. As shown in \cref{fig:teaser}\textcolor{red}{a}, the quality of the human mesh reconstruction is poor when only the legs are visible.

To address these problems, we propose a ``\methodf{}'' (\method{}) bottom-up human body mesh reconstruction approach (shown in \cref{fig:teaser}\textcolor{red}{b}). When only a few human body parts are visible, capturing the entire human body can be challenging for the network, but capturing the visible human body parts is more feasible. Besides, when there are only a few visible parts in the input, the interdependence of top-down approaches brings less knowledge and more noise to each other. If we do the reconstruction part by part, we can naturally avoid both information capturing and inter-part interference problems. However, there are also challenges of part-by-part reconstruction. Two main challenges are: a) independent reconstruction of all mesh parts and b) adjacent part fusion when more than one part is visible. To address the first challenge, we design a set of parametric models for each body part. Each parametric model can take a few shape parameters along with part global transformations as input to reconstruct the mesh of that body part without relying on other parts. For the second challenge, we design a fusion module to connect the adjacent visible parts together. We also design overlapping areas in the parametric models between two adjacent parts, which makes the fusion easier.

Specifically, our work is divided into the following steps. We first design and train the \pcaf{} (\pca{}). We generate the part templates using the template mesh of SMPL. The SMPL mesh is divided into 15 parts, and each part is trained using a large volume of SMPL ground truth. By regressing a few shape parameters and global transformations, we can obtain various shapes of each human part mesh. Second, we build up a network that takes monocular images as input and reconstructs human parts independently. We use a transformer-based backbone to get image features, and use them to regress the parameters of \pca{} to generate the part meshes. Then, we design a fusion module that connects the adjacent parts using a gradually-changed weighted-sum strategy, which could connect the mesh part seamlessly. Additionally, to 
evaluate our \methodf{} method, we constructed a benchmark comprised of images crops with partially visible human bodies and corresponding \pca{} annotations. 
We use existing public datasets Human36m and 3DPW to generate our benchmark \hmf{} and \pwf{}, and use both datasets for evaluation. In training, we use a similar image cropping strategy as augmentations on existing public datasets to get similar input image domains as in testing benchmarks, hence obtaining better performance.

To summarize, our contributions are as follows:
\begin{itemize}
    \item We designed a \methodf{} bottom-up solution that can reconstruct the human body mesh part by part. Different from SMPL-based top-down approaches, our method can independently reconstruct visible body parts when a large portion of the human body is not visible.
    \item We designed a set of parametric models representing every body part. Different from SMPL, 
    these parametric models can independently represent each body part, and be easily connected together when needed.
    \item We design a fusion module to smoothly connect each part together when multiple visible parts exist.
    \item In order to evaluate our method, we established 2 benchmark datasets with partially visible human image croppings and \pca{} annotations. We also design a similar augmentation strategy on the training dataset to improve our mesh reconstruction quality.
\end{itemize}

Our experiments validate the expressive ability of our parametric model and show that our reconstruction method outperforms the state-of-the-art on multiple datasets for partially visible human body input.

\section{Related Works}

\noindent
\textbf{Human body mesh reconstruction.}
Human body mesh reconstruction has been a popular research area.  Previous works~\cite{kanazawaHMR18,kanazawa2019learning,kocabas2020vibe,sun2019dsd-satn,zeng20203d,luan2021pc,luan2023high,gong2022self,gong2023progressive, zhai2023language, wang2022groupdancer, wang2024dancecamera3d} apply whole-body parametric model SMPL~\cite{SMPL:2015} for mesh reconstruction, and achieve good results when the whole body is visible. As top-down approaches, they rely on capturing whole-body information and their performance strongly degrades when the extraction of whole-body features fails.  
A number of past studies \cite{zhou_human_2021, choi_learning_2022, liu_votehmr_2021, zhang_object-occluded_2020, yuan_glamr_2022, khirodkar_occluded_2022, kocabas_pare_2021} have concentrated on the recovery of comprehensive human body meshes using occluded monocular inputs. These techniques strive to infer obscured parts based on visible body parts, which would place a stronger emphasis on capturing entire body information over the accuracy of each visible body part. Thus, when the entire body is not readily recognizable, these approaches tend to have results similar to non-occlusion approaches.
Other methods like those proposed by \cite{zhao_through-wall_2019, xue_mmmesh_2021, li_unsupervised_2022, khirodkar_occluded_2022, huang_occluded_2022} employ temporal inputs or other modalities (\eg, radio signals) to guide the reconstruction process, but their designs still focus on the body as a whole rather than individual parts. 
\cite{zuffi2015stitched} focused on a bottom-up mesh reconstruction strategy, but it is not designed for learning architectures and partially visible inputs. 
In this paper, we design a bottom-up learning-based approach that can tackle few-part-visible inputs.

\noindent
\textbf{Human pose estimation.}
Different from human mesh reconstruction, many human pose estimation approaches~\cite{simple,semanticsgcn,Ci_2019_ICCV,Cai_2019_ICCV,pavllo:videopose3d:2019,zhang2021learning} are using a bottom-up design as we do on mesh in this paper. They use a lifting strategy to estimate 3D body joints from 2D key points. However, inferring a complete 3D human mesh from a sparse set of 3D joints is an ill-posed and challenging task. Existing works would use 3D joints as a byproduct and supervision. Previous works \cite{cheng_occlusion_aware_2019, forsyth_multiple_2008, rafi_semantic_2015, radwan_monocular_2013, banik_occlusion_2023, zhou_human_2021, veges_temporal_2020, chen_multi-person_2023} can tackle slight occlusions, but they did not focus on the largely invisible cases. Since even 3D or 2D pose estimation would rely on inter-joint/key point correlation, the largely invisible case would compromise the basic 2D key points, resulting in unsatisfying results.

\section{\methodf{}}
\label{sec:method}
\subsection{Overview}
\noindent
\textbf{Problem formulation.}
In this paper, our task is to address the challenge of human mesh reconstruction from an image where only a limited portion of the human body is visible. Specifically, the input for this task is a monocular image $I$ containing a partially visible human body, and our objective is to develop a method $F(\cdot)$ capable of reconstructing the visible human body part mesh $v$ from the given image, \textit{a.k.a.}\ $v=F(I)$. In the following sections, we elaborate on how $F(\cdot)$ is designed.

\begin{figure*}[t]
    \centering
    \includegraphics[width=1\linewidth]{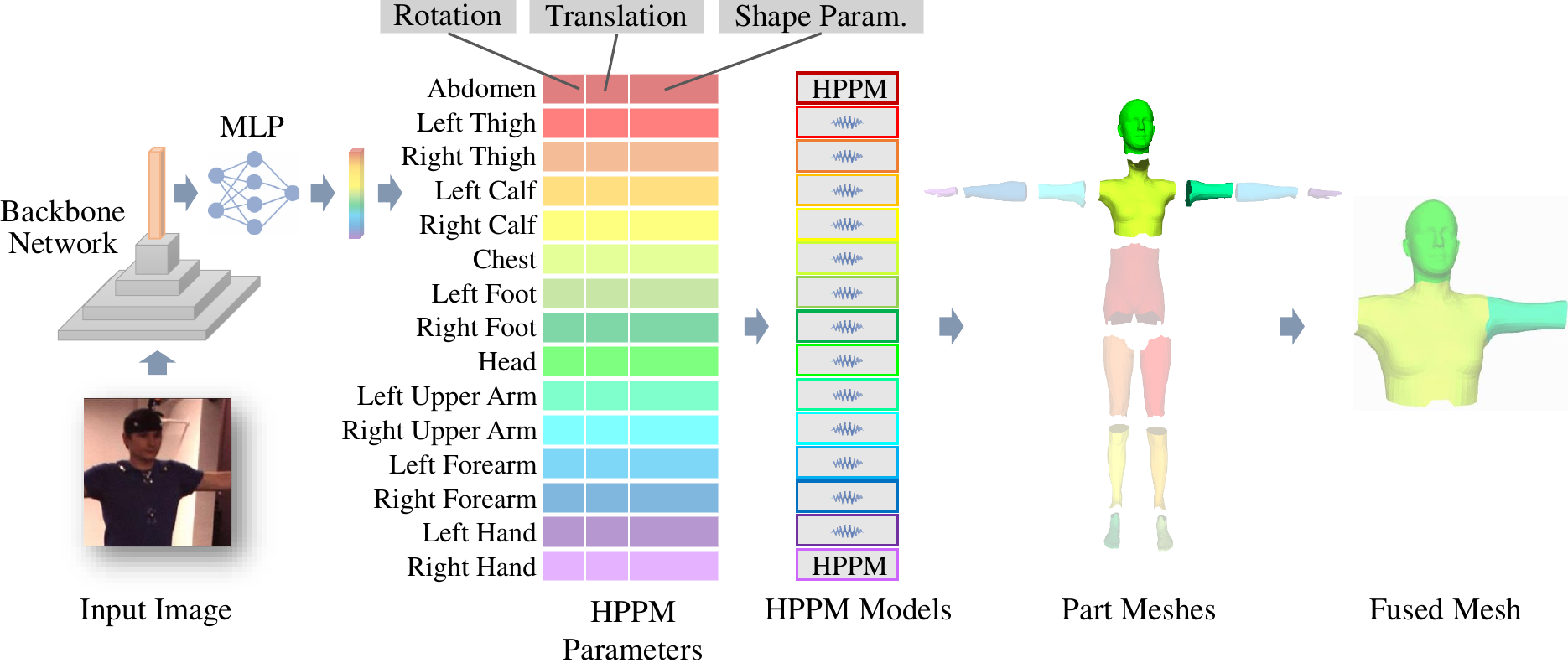}
    \caption{Our \methodf{} (\method{}) method takes a monocular partially visible human image as input and generates the human mesh of visible parts. The input image first goes through a backbone and an MLP network to get the parameters of \pca{}. Then, these parameters are used to generate part meshes through each part-specific \pca{}. Finally, a fusion module connects adjacent visible parts. Details are provided in \cref{sec:method}.} 
    \label{fig:pipeline}
\end{figure*}

\noindent
\textbf{Framework overview.}
As illustrated in \cref{fig:pipeline}, our pipeline starts with the extraction of image features through a backbone network, followed by the use of a multi-layer perceptron (MLP) to obtain the features required by the \pcaf{} (\pca{}). In \pca{}, we divide the human body into 15 parts. For each part, \pca{} requires 3 inputs: translation, rotation, and shape parameters. Upon the shape adjustments made by the shape parameters, \pca{} employs simple rigid transformations to generate the mesh of each body part without explicitly incorporating the human pose. 
\pca{} outputs body part meshes and joints, and is supervised by corresponding ground truths via multiple loss functions, ensuring that each mesh part can be reconstructed independently. 
Moreover, a fusion module is proposed that seamlessly integrates multiple visible mesh parts. In \cref{sec:pca}, we introduce the design and training of the \pca{}. \cref{sec:independent} elaborates on the design of our independent reconstruction network, and \cref{sec:fusion} introduces our fusion module.

\subsection{\pcaf{}}
\label{sec:pca}

\begin{figure*}[t]
    \centering
    \includegraphics[width=1\linewidth]{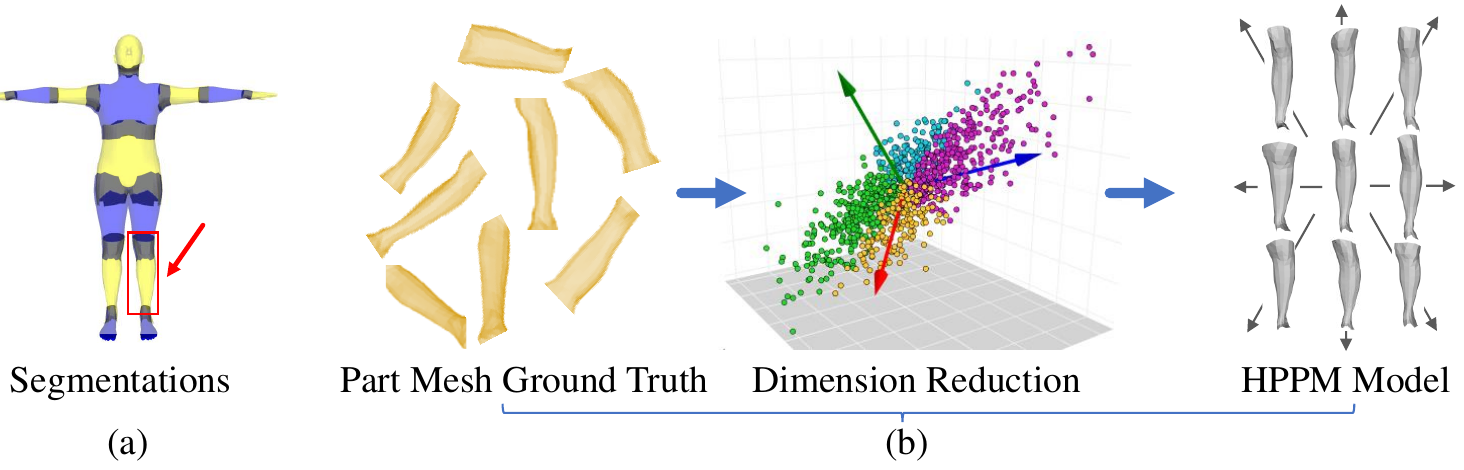}
    \caption{\textbf{(a)} \pca{} template segmentation. We segment the SMPL template to generate \pca{} templates. The joint areas are covered by both adjacent parts (overlap). This design allows \pca{} to naturally cover near-joint distortions using shape parameters, while also facilitating the fusion of parts together. \textbf{(b)} \pca{} training process. We segment part ground truths from Human3.6M~\cite{human36m}, 3DPW~\cite{pw3d}, and AMASS~\cite{AMASS:2019}. For each part, we use a dimension-reduction strategy to train a matrix that maps the high-dimensional part meshes into a few shape parameters. Shape parameters are estimated by the network to recover part meshes.}
    \label{fig:pca}
\end{figure*}

We design \pcaf{} (\pca{}) to facilitate the independent reconstruction of each body part. Unlike the widely-used holistic approach SMPL, \pca{} allows for the decoupling of body parts, eliminating the reliance on inter-part correlations within the model, \ie, to improve performance when the inter-part correlation is not stable in input images. The \pca{} design consists of two stages. First, a mesh template is defined for each body part; then a linear function is trained to the shape parameters into part meshes.

\noindent
\textbf{Part template design.} We craft part-specific template meshes utilizing the template mesh from SMPL. We segment this mesh into 15 distinct parts based on the blend weight parameter $W$ of SMPL, where each entry of blend weight matrix $W_{ij}$ indicates how vertex $v_i$ is influenced by bone $b_j$ when the limbs move. Here, we first group the vertices using the following strategy:
\begin{equation}
    p_i = \mathop{\arg \max}_{j}{W_{ij}},
\end{equation}
where $p_i$ is the part index of vertex $v_i$.
\Ie, we first assign each vertex to the bone with the highest blend-shape contribution.
This grouping strategy ensures that vertices assigned to the same section are attached to the same body limb, allowing us to treat each part as a nearly rigid mesh, so that the model would not need explicit inner part poses. 
Obtaining the grouping as raw segmentation, we manually fuse some raw segments to create a cohesive set of 15 parts. This strategy minimizes unnecessary divisions so that body parts that are empirically considered nearly rigid when put together will be combined into a single part. \Eg, we combine the shoulders and neck with the torso section as they collectively form a nearly rigid structure. However, we do not merge the thigh and calf to be a single part because the knee joint allows significant bending.
Finally, we perform dilation for each part to create some overlapping between adjacent parts. This design not only incorporates near-joint mesh deformation into the deformation of each part, but also simplifies the fusion between adjacent parts. More intricate details of \pca{} design are further explained in \supl{A}. The final part templates are visualized in \cref{fig:pca}\textcolor{red}{a}.

\noindent
\textbf{Parameter-adjustable \pca{} training.}
Having the template of each part, we train our \pca{} on Human3.6M~\cite{human36m}, 3DPW~\cite{pw3d}, and AMASS~\cite{AMASS:2019} datasets. Our training process is visualized in \cref{fig:pca}\textcolor{red}{b}. First, we segment the ground truth part meshes using the template part mesh vertex indices obtained above. Then we use a dimension reduction method to train our linear mapping. For part $p$, we form the mesh data in the training datasets into a matrix $X_p \in \mathbb{R} ^{n \times m}$, where $m$ is the number of samples and $n=3N$ is the number of vertices times their dimension. We reduce
the matrix dimensionality from $n$ to $k$ using principal component analysis. We denote as $\mathcal{U}_p \in \mathbb{R}^{k\times n}$ the matrix that maps the $k$ dimensional shape parameters to part meshes. 
We use an adjustable number of parameters for each part, so that we can adapt the fitting capacity of different parts given different shape variances.
As demonstrated in \cref{fig:pcanpara}\textcolor{red}{a}, we show the relationship between fitting accuracy and parameter dimensions for each part of the training set. 
Here, we set a maximum joint and vertex fitting error for each part while maintaining a minimal dimension. We empirically set the maximum error allowance for both joint and vertex to 2mm and the minimum dimension to 16. In \cref{tab:hppm_ds} we show the training errors of each part. The experiment in \cref{tab:abla} ``w/ fixed \#parameters'' row reveals that our model, with a total parameter count of 360, outperforms the method that uses a fixed dimension of 24 per part, which has the same total number of parameters. 

\noindent
\textbf{Joint regressor.} \pca{} is also designed to generate body joints so that 3D joints can be estimated along with part meshes. We train a joint regressor matrix for each body part as 
\begin{equation}
    {\mathcal{J}_p} = \mathop{\arg\min}_{\mathcal{J}_p^{\prime}}\|J - {\mathcal{J}_p^{\prime}}v_{0}\|_F,
\end{equation}
where $J$ denotes the ground truth joints, and $\|\cdot\|_F$ is the Frobenius normal. Note that for every part, we only regress the joint close to that part. The part-joint correspondence is shown in \supl{B}.

Our \pca{} model is finally defined as $\{\mathcal{U}_p, \mathcal{J}_p\}$, with $p$ for the $p$-th part. Compared with SMPL, \pca{} is capable of determining a human part mesh independently through overall translation, rotation, and a limited set of deformation parameters. This design enables precise reconstruction of body segments.

\subsection{Divide: Part Independent Reconstruction}
The divide stage provides a solution for reconstructing each part independently from images. Input images are processed through a Swin Transformer \cite{liu2021swin} backbone to obtain image features, which are then passed through a multi-layer perceptron (MLP) model to derive the HPPM shape parameters $\hat{S}_p$, global rotation $\hat{R}_p$, and global translation $\hat{T}_p$ for part $p$. As multiple rotations are being estimated, our network outputs 6D rotation~\cite{zhou2019continuity} to improve convergence on rotation estimation. We write the global transformation in one matrix $\hat{M}_p$ and obtain the estimated mesh for part $p$ as:
\label{sec:independent}
\begin{equation}
    \hat{v}_p = \hat{M}_p(\mathcal{U}_p\hat{S}_p + \mathcal{M}_p), \;\text{ with } \hat{M}_p = \begin{pmatrix}
    \hat{R}_p & \hat{T}_p \\
    \textbf{0} & 1 \\
    \end{pmatrix}, 
\end{equation}
where $\mathcal{U}_p$ and $\mathcal{M}_p$ is the shape matrix and mean shape in \pca{}, and $\hat{v}_p$ is in homogeneous coordinates. 
From these results,
we calculate the part joints using regressors:
\begin{equation}
    \hat{J}_p = \mathcal{J}_p\hat{v}_p,
\end{equation}
thus inferring each part mesh and corresponding joint independently.

\noindent
\textbf{Training losses.} We design the following losses to ensure the independent reconstruction of every part. In order to achieve part independence, all the supervisions in this section are defined on body parts.

First, we use the mesh and joint annotation of each part to directly supervise part mesh and joint. Specifically, we define part vertex loss $\mathcal{L}_v$ and part 3D joint location loss $\mathcal{L}_{j3d}$ as:
\begin{equation}
\begin{array}{cc}
    &\mathcal{L}_v = \sum_{p}\sum_{i}\delta_p\|\hat{v}_{pi}-v_{pi}\|, \\
    &\mathcal{L}_{j3d} = \sum_{p}\sum_{i}\delta_p\|\hat{J}_{pi}-J_{pi}\|,
\end{array}
\label{eq:loss_v}
\end{equation}
where $\hat{v}_{pi}$ and $v_{pi}$ are the $i$-th vertex of the $p$-th estimated part mesh and ground truth part mesh, respectively. $\hat{J}_{pi}$ and $J_{pi}$ are estimated 3D joint location and ground truth joint location, respectively. $\delta_p$ indicates the visibility. $\delta_p=1$ when that part is visible and $\delta_p=0$ when its not. $\|\cdot\|$ indicates the L2-norm of a vector. 
Besides mesh and 3D joint loss, we also include 2D joint projection loss to enhance 2D projection accuracy as
\begin{equation}
    \mathcal{L}_{j2d} = \sum_{p}\sum_{i}\delta_p\|\Pi\hat{J}_{pi}-\Pi J_{pi}\|,
\end{equation}
where $\Pi$ is the projection matrix from camera to image coordinate.

Apart from the above losses, we also directly supervise the part parameters and their global transformations. Specifically, we define part shape parameter loss $\mathcal{L}_{s}$, part rotation loss $\mathcal{L}_{r}$, and part translation loss $\mathcal{L}_{t}$ as
\begin{equation}
    \begin{array}{cc}
    \mathcal{L}_{s} &= \sum_{p}\delta_p\|\hat{S}_p-S_p\|, \\
    \mathcal{L}_{r} &= \sum_{p}\delta_p\|\hat{R}_p-R_p\|, \\
    \mathcal{L}_{t} &= \sum_{p}\delta_p\|\hat{T}_p-T_p\|,
    \end{array}
\end{equation}
where $\hat{S}_p$, $S_p$, $\hat{R}_p$, $R_p$, $\hat{T}_p$, and $T_p$ are defined similar to \cref{eq:loss_v}.

In total, the loss for part independent reconstruction is defined as:
\begin{equation}
    \mathcal{L}_{div} = \lambda_{v}\mathcal{L}_{v} + \lambda_{j3d}\mathcal{L}_{j3d} + \lambda_{j2d}\mathcal{L}_{j2d} + \lambda_{s}\mathcal{L}_{s} + \lambda_{r}\mathcal{L}_{r} + \lambda_{t}\mathcal{L}_{t},
\end{equation}
where $\lambda_{v}$, $\lambda_{j3d}$, $\lambda_{j2d}$, $\lambda_{s}$, $\lambda_{r}$, and $\lambda_{t}$ are loss weights.

\subsection{Fuse: Adjacent Part Fusion}
\label{sec:fusion}
The fuse part is designed to combine the visible part meshes into a single mesh when multiple parts are visible. Before applying the connection, we use two self-supervision fusion losses---namely overlapping loss and depth consistency loss---to bring the part meshes closer to each other. The overlapping loss is defined:
\begin{equation}
    \mathcal{L}_{ol} = \sum_{p}\sum_{v \in O_p}\delta_p\|\hat{v}_p-\bar{v}\|,
\end{equation}
where $O_p$ is the vertex set that is in the overlapping area between 2 adjacent parts, $\delta_p$ is defined the same as in \cref{eq:loss_v}, and $\bar{v}$ is the average vertex location of the overlapping area of 2 adjacent parts. It is computed as:
\begin{equation}
    \bar{v}=\frac{\sum_{v \in O_{p} \cap O_{p^{\prime}}}\hat{v}_p}{2|{O_{p} \cap O_{p^{\prime}}}|},
\end{equation}
where $O_{p^{\prime}}$ is the overlapping area on the adjacent part mesh.
This way, we can ensure that the junction vertices of adjacent parts are both closer to their average, thereby closer to each other. The depth consistency loss is designed to constrain the parts that occur in the same input image but are not directly connected. A regularization term is applied to those parts as:
\begin{equation}
    \mathcal{L}_{dc} = \sum_{p}\sum_{i}\delta_p\|\hat{v}^{z}_{pi}-\bar{v}^{z}\|,
\end{equation}
where $\hat{v}^{z}_{pi}$ is the $z$ coordinate of the $i$ vertex location of $p$-th estimated part, and $\bar{v}^{z} = \frac{\sum_{p}\sum_{i}\hat{v}^{z}_{ip}}{\sum_{p}N_p}$ is the average $z$ coordinate of all vertices in all visible parts. 

Then our self-supervised fusion loss is defined as:
\begin{equation}
    \mathcal{L}_{fu} = \lambda_{ol}\mathcal{L}_{ol} + \lambda_{dc}\mathcal{L}_{dc},
\end{equation}
where $\lambda_{ol}$, $\lambda_{dc}$ are loss weights. Our total loss function is: 
\begin{equation}
    \mathcal{L} = \mathcal{L}_{div} + \mathcal{L}_{fu}.
\end{equation}

\noindent
\textbf{Gradual part connecting.} Besides the aforementioned training supervision, we also design a post-processing connecting module to seamlessly attach adjacent parts together during inference, based on a weighted-sum strategy.
We identify two types of output vertices $v^{c}_{k}$ when connecting two adjacent parts into one mesh: the final vertices that belong to the overlapping region shared by both part meshes, and those that do not belong to this region. 
In non-overlapping regions, the final vertices are computed:  
\begin{equation}
    v^{c}_{k} = \hat{v}_{pi},
\end{equation}
where $\hat{v}_{pi}$ is the corresponding vertex of $v^{c}_{k}$ in part $p$. Here, ``corresponding'' means the $i$-th vertex in part $p$ is topologically the $k$-th vertex in the SMPL template. 
If $v^{c}_{k}$ is in the overlapping area, then:
\begin{equation}
    v^{c}_{k} = v^{c}_{p_1i} = v^{c}_{p_2j} = \frac{\hat{v}_{p_1i} d_{2j} + \hat{v}_{p_2i} d_{1i}}{d_{1i} + d_{2j}}.
\end{equation}
Here, $d_{1i}$ is the shortest topology distance from $\hat{v}_{p_1i}$ to the nearest non-overlapping vertex in $p_1$th part mesh. That means, if $d_{1i}=2$, $\hat{v}_{p_1i}$ needs to go through one other vertex to connect to the nearest non-overlapping vertex in the $p_1$th part mesh. $d_{2j}$ is defined similar to $d_{1i}$.
In this fusion process, the overlapping loss is used to ensure the vertices in the overlapping areas are closely aligned, and the gradual part connecting process can further eliminate the small vertex difference. We use this design to avoid undesirable deformations in the fused mesh.

\begin{table}[t]
    \centering
    \caption{Per-part parameters and registration errors. 
    For each part $p$, we first provide their number of shape parameters $k_p$, number of vertices $N_p$, and number of body joints $|\mathcal{J}_p|$, followed by their mean $\ell_2$ errors (in mm) \wrt GT vertices and joints. For both joint and vertex errors, we report the training error in the HPPM optimization process (\textcolor{red!60}{red} boxes), along with the error in the ground truth fitting process for both datasets (\textcolor{blue!60}{blue} boxes for \hm{} and \textcolor{cyan!90}{cyan} boxes for \pw{}). 
    }
    \resizebox{\linewidth}{!}{
    \begin{tabular}{l|ccc|>{\columncolor{red!15}}c>{\columncolor{blue!15}}c>{\columncolor{cyan!15}}c|>{\columncolor{red!15}}c>{\columncolor{blue!15}}c>{\columncolor{cyan!15}}c}
    \hline
        \multirow{2}{*}{Part Names} & \multicolumn{3}{c|}{Hyperparameters} & \multicolumn{3}{c|}{Vertex Errors} & \multicolumn{3}{c}{Joint Errors} \\
        \cline{2-10}
        
        & $\;\, k_p\quad$ & \multicolumn{1}{c}{$N_p$} & $|\mathcal{J}_p|$ & Train. & Fit. {\scriptsize PV-H36M} & Fit. {\scriptsize PV-3DPW} & Train. & Fit. {\scriptsize PV-H36M} & Fit. {\scriptsize PV-3DPW} \\
        \hline

        Abdomen         & 33 & 839  & 4  & 1.95 & 1.00 & 1.81 & 1.67 & 1.06 & 1.22 \\
Left Thigh      & 31 & 375   & 2  & 1.29 & 0.86 & 0.98 & 2.00 & 1.81 & 2.13 \\
Right Thigh     & 31 & 370   & 2  & 1.25 & 0.87 & 0.92 & 2.00 & 0.96 & 1.93 \\
Left Calf       & 16 & 284   & 2  & 1.07 & 0.62 & 0.92 & 0.79 & 0.70 & 0.24 \\
Right Calf      & 16 & 284   & 2  & 1.10 & 0.67 & 0.85 & 1.23 & 1.42 & 1.25 \\
Chest           & 42 & 1,490 & 4  & 2.00 & 0.90 & 2.09 & 0.95 & 0.47 & 0.93 \\
Left Foot       & 16 & 283   & 1  & 0.53 & 0.27 & 0.45 & 0.48 & 0.57 & 1.78 \\
Right Foot      & 16 & 283   & 1  & 0.57 & 0.28 & 0.41 & 0.87 & 1.32 & 1.28 \\
Head            & 21 & 1,257 & 3  & 0.59 & 0.45 & 0.50 & 1.89 & 1.96 & 4.34 \\
Left Upper Arm  & 36 & 381   & 2  & 0.81 & 0.37 & 0.64 & 1.96 & 1.08 & 0.80 \\
Right Upper Arm & 38 & 382   & 2  & 0.80 & 0.33 & 0.84 & 1.90 & 0.69 & 3.21 \\
Left Forearm    & 16 & 316   & 2  & 1.08 & 0.52 & 0.83 & 1.13 & 0.66 & 1.19 \\
Right Forearm   & 16 & 316   & 2  & 1.11 & 0.55 & 0.97 & 1.37 & 0.95 & 1.75 \\
Left Hand       & 16 & 810   & 1  & 0.57 & 0.29 & 0.60 & 1.34 & 0.77 & 2.24 \\
Right Hand      & 16 & 810   & 1  & 0.56 & 0.32 & 0.69 & 1.52 & 1.24 & 0.46 \\
\hline
Average         & - &  -  & - & 1.11 & 0.59 & 1.05 & 1.46 & 1.05 & 1.69 \\
    \hline
    \end{tabular}
    }

    \label{tab:hppm_ds}
\end{table}

\begin{table*}[t]
    \centering
    \caption{Comparison of \method{} with recent previous approaches on our \hmf{} and \pwf{} benchmarks. MPVE and MPJPE are in millimeters, lower means better mesh and joint accuracy. The best results are shown in \textbf{bold}. The left 2 columns are the results directly tested on our benchmark using released model weights, among which ``D\&F (Ours)'' are trained on public datasets. The 2 columns on the right are the results finetuned using our partially visible augmentation and part mesh pseudo ground truth. The results show our method outperforms recent previous methods, and our partially visible augmentation and part mesh pseudo ground truth can contribute to the improvement on partially visible human image inputs. 
    }
    \resizebox{\linewidth}{!}{
    \begin{tabular}{l||>{\columncolor{blue!15}}c>{\columncolor{blue!15}}c|>{\columncolor{cyan!15}}c>{\columncolor{cyan!15}}c||>{\columncolor{blue!15}}c>{\columncolor{blue!15}}c|>{\columncolor{cyan!15}}c>{\columncolor{cyan!15}}c}
    \hline
        \multirow{3}{*}{Methods} & \multicolumn{4}{c||}{Directly Tested} & \multicolumn{4}{c}{Finetuned on Our Augmentation} \\ \cline{2-9}
        & \multicolumn{2}{c|}{\cellcolor{blue!15}\hm{}} & \multicolumn{2}{c||}{\cellcolor{cyan!15}\pw{}} & \multicolumn{2}{c|}{\cellcolor{blue!15}\hm{}} & \multicolumn{2}{c}{\cellcolor{cyan!15}\pw{}} \\ \cline{2-9}
        & MPVE$\downarrow$ & MPJPE$\downarrow$ & MPVE$\downarrow$ & MPJPE$\downarrow$ & MPVE$\downarrow$ & MPJPE$\downarrow$ & MPVE$\downarrow$ & MPJPE$\downarrow$ \\ \hline
        
        MotionBERT~\cite{motionbert2022} & 350.6 & 305.4 & 284.7 & 250.0 & 196.9 & 169.6 & 185.5 & 155.4 \\ 
        SEFD~\cite{yang2023sefd} & 336.7 & 259.6 & 275.0 & 224.6 & 276.0 & 198.9 & 241.1 & 203.0 \\ 
        GLoT~\cite{shen2023global} & 325.2 & 298.5 &  297.2 & 255.7 & 214.1 & 199.0 & 235.0 & 213.5 \\ 
        CycleAdapt~\cite{nam2023cyclic} & 367.0 & 318.1 & 268.2 & 230.8 & 249.4 & 231.9 & 189.0 & 137.1 \\ 
         \hline
        \method{} (Ours) & \textbf{155.7} & \textbf{156.3} & \textbf{208.6} & \textbf{194.9} &  \textbf{63.3} & \textbf{55.9} & \textbf{109.9} & \textbf{102.7} \\ \hline
    \end{tabular}
    }

    \label{tab:sota}
    \vspace{-0.2cm}
\end{table*}

\begin{table*}[t]
    \centering
    \caption{We do ablation studies on designing necessity and loss functions on \hm{} and \pw{} benchmarks. As we remove some losses and module designs, the performance suffers decreases to different extents. The ``Directly Tested'' result of \method{} is when we train on public datasets and test on our partially visible benchmarks. The best results are shown in \textbf{bold}.}
    \resizebox{\linewidth}{!}{
    \begin{tabular}{l|>{\columncolor{blue!15}}c>{\columncolor{blue!15}}c|>{\columncolor{cyan!15}}c>{\columncolor{cyan!15}}c}
    \hline
        \multirow{2}{*}{Experiment settings} & \multicolumn{2}{c|}{\cellcolor{blue!15}\hm{}} & \multicolumn{2}{c}{\cellcolor{cyan!15}\pw{}} \\ 
        \cline{2-5}
        & MPVE/mm$\downarrow$  & MPJPE/mm$\downarrow$ & MPVE/mm$\downarrow$ & MPJPE/mm$\downarrow$ \\ 
        \hline
        w/o part 2D projection loss & 64.2 & 56.7 & 111.8 & 104.9 \\ 
        w/o part 3D joint loss & 63.9 & 56.2 & 112.4 & 105.8 \\ 
        w/o part 3D per-vertex loss & 68.4 & 63.5 & 120.2 & 108.3 \\ 
        w/o \pca{} shape-parameter loss & 70.1 & 57.0 & 119.7 & 103.9\\ 
        w/o \pca{} 6D rotation loss & 95.6 & 87.5 & 138.5 & 127.4 \\ 
        w/o \pca{} translation loss & 75.1 & 69.9& 123.0 & 115.3 \\ 
        w/o overlapping loss & 74.5 & 64.2 & 125.2 & 111.8\\ 
        w/o depth consistency loss & 65.4 & 58.5 & 113.9 & 104.1\\
        w/ fixed \#parameters & 67.5 & 61.7 & 114.0 & 105.4\\
        \hline 
        \method{}(Ours) & \textbf{63.3} & \textbf{55.9} & \textbf{109.9} & \textbf{102.7}\\ \hline
    \end{tabular}
    }

    \label{tab:abla}
\end{table*}

\begin{table}[t]
    \centering
    \caption{Results on \pw{} images with more visible parts. The best results are shown in \textbf{bold}. As the number of visible parts increases to 5-10, our method still works well and outperforms previous approaches.}
    \resizebox{0.75\linewidth}{!}{
    \begin{tabular}{l|cc|cc}
    \hline
        \#Parts Visible & \multicolumn{2}{c|}{1-4} & \multicolumn{2}{c}{5-10} \\ \hline 
        \hline
        Methods & MPVE/mm$\downarrow$ & MPJPE/mm$\downarrow$ & MPVE/mm$\downarrow$ & MPJPE/mm$\downarrow$  \\ \hline 
        CycleAdapt~\cite{nam2023cyclic} & 189.0 & 137.1 & 169.9 & 132.8 \\ 
        GLoT~\cite{shen2023global}& 185.5 & 155.4 & 142.3 & 131.9 \\ 
        \method{} (Ours) & \textbf{109.9} & \textbf{102.7} & \textbf{117.4} & \textbf{107.5} \\ \hline
    \end{tabular}
    }
    \label{tab:better}
    \vspace{-0.5em}
\end{table}

\begin{figure*}[t]
    \centering
    \includegraphics[width=1\linewidth]{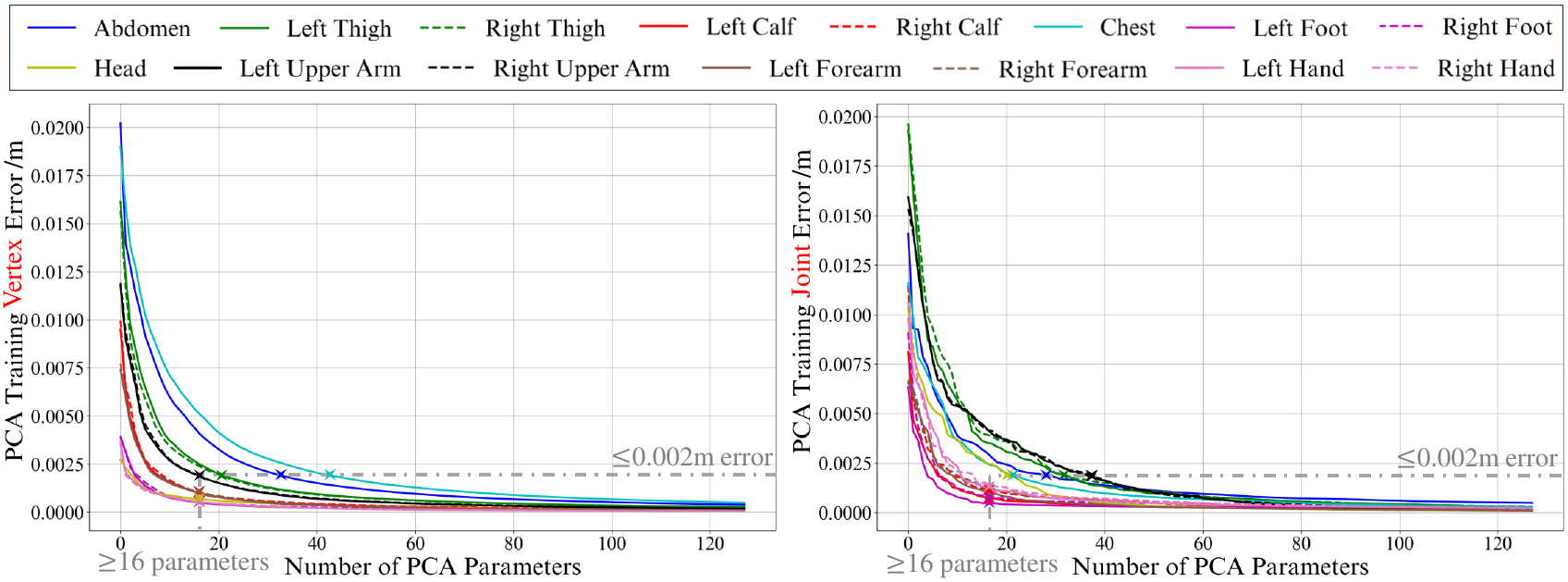}
    \caption{\pca{} training error of each part changes with the number of shape parameters used. We consider an adjustable number of parameters for each part. We set the maximum joint and vertex training errors to be 2mm, and a minimum number of parameters to 16. Left: vertex training error. Right: joint training error.}
    \label{fig:pcanpara}
\end{figure*}

\begin{figure}[t]
    \centering
    \includegraphics[width=1\linewidth]{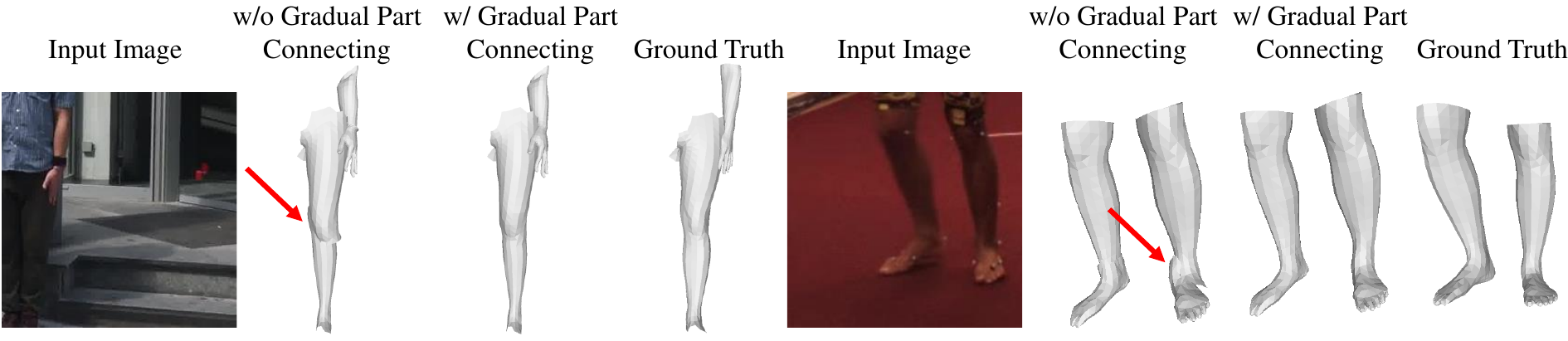}
    \caption{Visual ablation on gradual part connecting. When this module is removed, the connection points between two adjacent parts become misaligned, as indicated by the \textcolor{red}{red} arrow. This alignment issue is resolved using the gradual part connecting.}
    \label{fig:abla}
    \vspace{-0.2cm}
\end{figure}

\begin{figure*}[t]
    \centering
    \includegraphics[width=1\linewidth]{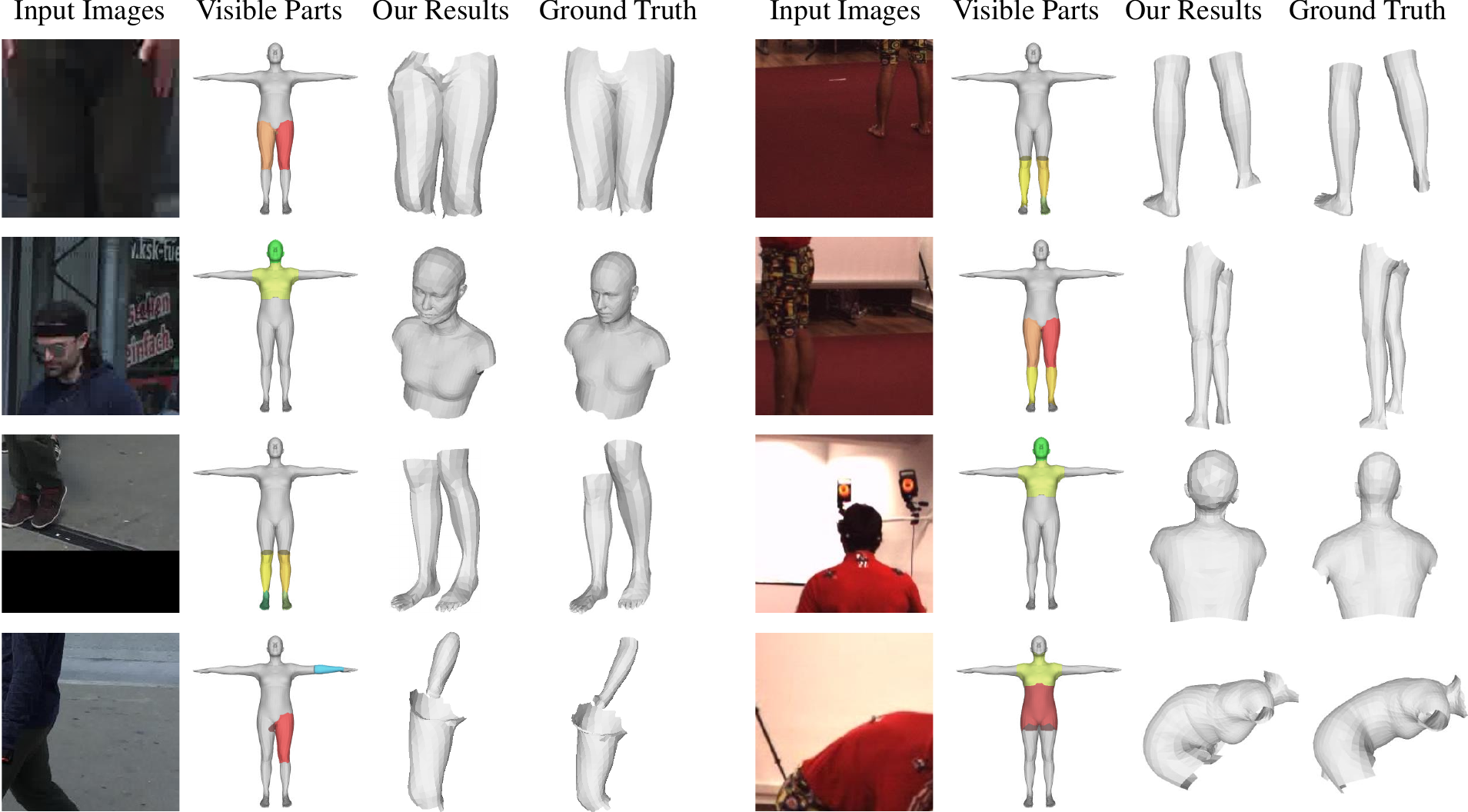}
    \caption{Qualitative results from \pw{} benchmark (left) and \hm{} benchmark (right). Each benchmark features an input image (\nth{1} column), selected visible parts (\nth{2} column), our mesh results (\nth{3} column), and the ground truth mesh for visible parts only (\nth{4} column). Our method successfully generates accurate meshes from partially visible inputs.}
    \label{fig:vis}
\end{figure*}

\section{Experiments}
\subsection{\dsf{} Benchmarks}
\label{sec:benchmark}

To train and evaluate our approach, we created two benchmarks, \hmf{}(\hm{}) and \pwf{}(\pw{}) based on existing public datasets, Human3.6M~\cite{human36m} and 3DPW~\cite{pw3d}, respectively. The input images for our benchmarks are partially visible human images. Additionally, we generated the corresponding \pca{} annotations. In our experiments, we utilize the mean per-vertex error (MPVE) to assess the accuracy of the mesh reconstruction and the mean per-joint position error (MPJPE) to evaluate the precision of the joint positions, following~\cite{kanazawaHMR18}. The detailed definition of MPVE and MPJPE can be found in \supl{C}.

\noindent
\textbf{\pca{} annotations.} We generate \pca{} annotations from SMPL ground truth of Human3.6M and 3DPW. The annotations consist of shape parameters $S$ and global transformation $M$, which are used as annotations for training. From here and the following representations in this section, we omit the part subscript $p$ for simplicity. Specifically, we first use the segmentation strategy in \cref{sec:pca} to generate ground truth part meshes. For each ground truth body part mesh $v$ and its corresponding part template $v_0$, we calculate the global transformation $M$ from $v_0$ to $v$ as:
\begin{equation}
    M = \mathop{\arg\min}_{M{\prime}}\|v - M^{\prime}v_{0}\|,
    \label{eq:annot_g}
\end{equation}
where $v$ and $v_0$ are in homogeneous coordinates and $M \in \mathbb{R}^{4 \times 4}$. In practice, we use the least-square method to solve \cref{eq:annot_g}. Given the global transformation, we can transform ground truth mesh $v$ to the canonical space of template mesh as $v^{\prime} = M^{\top}v$. Then we calculate the \pca{} shape parameter $S$ as:
\begin{equation}
    S = \mathcal{U}^{\top}(v^{\prime} - \mathcal{M}),
\end{equation}
where $\mathcal{U}^{\top}$ is the transpose matrix of \pca{} shape matrix, $\mathcal{M}$ is the mean shape of \pca{}, and $v^{\prime} \in {\mathbb{R}^{3N}}$ is the flattened vertex vector, given $N$ the vertex number of this part. Having shape parameters and global transformations, we can recover the part mesh as:
\begin{equation}
    v_{\textit{\pca{}}} = M(\mathcal{U}S + \mathcal{M}),
\end{equation}
In \cref{tab:hppm_ds}, we evaluate the \pca{} annotation error for each part in both benchmarks. We observe that for all body parts, the error is negligible and acceptable. We also evaluate the necessity of the generated supervisions in \cref{tab:abla}. The experiments show that the global transformation and shape parameter annotations are helpful to improve performance.

\noindent
\textbf{Partially visible human images.} We use a random cropping strategy to generate partially visible human images from Human3.6M and 3DPW. First, we project the \pca{} ground truth to the image, to determine which part of the image corresponds to which body part(s). Next, a center point is randomly selected within the human bounding box, as well as a random side length for the square cropping. For every image in the original dataset, this process is performed 20 times, and cropped results with 1 to 4 body parts visible are short-listed. For visible parts, we require $\geq 50\%$ area of the part bounding box to be inside the cropped image. Resulting cropped images are shown in \cref{fig:teaser} and \cref{fig:vis}.

\noindent
\textbf{Partially visible augmentation.} We use the similar image cropping strategy mentioned in the previous paragraph as training augmentation. This operation improves the generalizability of trained methods \wrt visibility distributions.
Our experiment in \cref{tab:sota} shows that our partially visible augmentation strategy not only increases the performance of our framework but also other previous methods on our partially visible datasets.

\subsection{Implementation Details}
We end-to-end train our network on a single NVIDIA A100 40GB GPU. We optimize our model on Human3.6M~\cite{human36m}, 3DPW~\cite{pw3d}, and SURREAL~\cite{varol17_surreal}, with the augmentation introduced in \cref{sec:benchmark}. We evaluate our model on the \hm{} and \pw{} benchmarks separately. The batch size is set to 92. The training weights are set to $\lambda_{v}=2.5$, $\lambda_{j3d}=1,250$, $\lambda_{j2d}=2,500$, $\lambda_{s}=100$, $\lambda_{r}=200$,  $\lambda_{t}=500$, $\lambda_{ol}=100$, and $\lambda_{dc}=1$. We use Adam~\cite{kingma2014adam} for optimization and the learning rate is set to $1\times 10^{-4}$. During inference, the dataset annotations for part visibility are leveraged. The code is implemented using PyTorch~\cite{paszke2019pytorch}.

\subsection{Results}

\noindent
\textbf{Comparison with state-of-the-art.}
We compare our method with recent previous methods on our \hm{} and \pw{} benchmarks on visible parts in \cref{tab:sota}. 
Our \method{} outperforms recent previous methods on both benchmarks in terms of both mesh and joint accuracy (MPVE/MPJPE), regardless of whether they are finetuned with our partially visible augmentation.

\noindent
\textbf{\pca{} parameter numbers.} To determine the number of shape parameters used in each part, \cref{fig:pcanpara} highlights how \pca{} training error \wrt part vertices and joints are impacted by the number of shape parameters. \Ie, the error drops as the number of parameters increases. The larger the number of parameters, the more exact the part meshes are, but the more challenging the prediction task becomes for \method{}. Therefore, we propose the trade-off by setting the maximum joint and vertex training errors to be 2 millimeters, and a minimum number of parameters to 16 in \pca{} design. In total, we use 360 shape parameters.

\noindent
\textbf{Ablation studies.}
We show the necessity of our module and loss function design in \cref{tab:abla}. The performance of our method drops when removing some necessary modules or loss functions. We observe that the part 3D per-vertex/3D joint/2D projection loss generally helps part mesh recovery and part joint accuracy, which is similar to the whole body SMPL-based frameworks. The HPPM pseudo-ground-truth annotations including 6D rotations are also helpful. The overlapping loss also increases the result by a large margin. Apart from these ablation studies, We also show the effectiveness of our gradual part connecting in \cref{tab:abla}. Without this module, the connection vertices between 2 adjacent parts can be misaligned (\textcolor{red}{red} arrow); an issue solved with the proposed connecting scheme. 

\noindent
\textbf{Higher-visibility scenarios.}
We show some results of our method when the input contains a higher number of visible parts in \cref{tab:better}. On \pw{} dataset, we increase the number of visible parts to 5-10. We observe that, even when there are more visible portions of the input human body, our method still works well and outperforms previous approaches.

\noindent
\textbf{Visualizations.}
We show some additional qualitative results on \hm{} and \pw{} benchmarks in \cref{fig:vis}. We observe that our method can generate valid mesh results with partially visible human image inputs.

\section{Conclusion}
In conclusion, our \methodf{} method successfully addresses the limitations of existing top-down human mesh reconstruction techniques in the presence of occlusions. Through \pcaf{}, independent part reconstruction, and strategic fusion, our approach consistently delivers accurate meshes with partially visible bodies, as validated by our provided benchmarks \hm{} and \pw{}. These advancements represent a considerable improvement in mesh reconstruction accuracy and reliability when dealing with partially visible human images.

\appendix
\section*{Supplementary Materials}

This supplementary material includes the following contents:
\begin{enumerate}[A.]
    \item \pca{} template design details.
    \item \pca{} part-joint correspondence.
    \item Evaluation metrics.
    \item Adaptiveness on full-body visible inputs.
    \item Future Works
\end{enumerate}
\section{\pca{} Template Design Details}
\begin{figure*}
    \centering
    \includegraphics[width=1\linewidth]{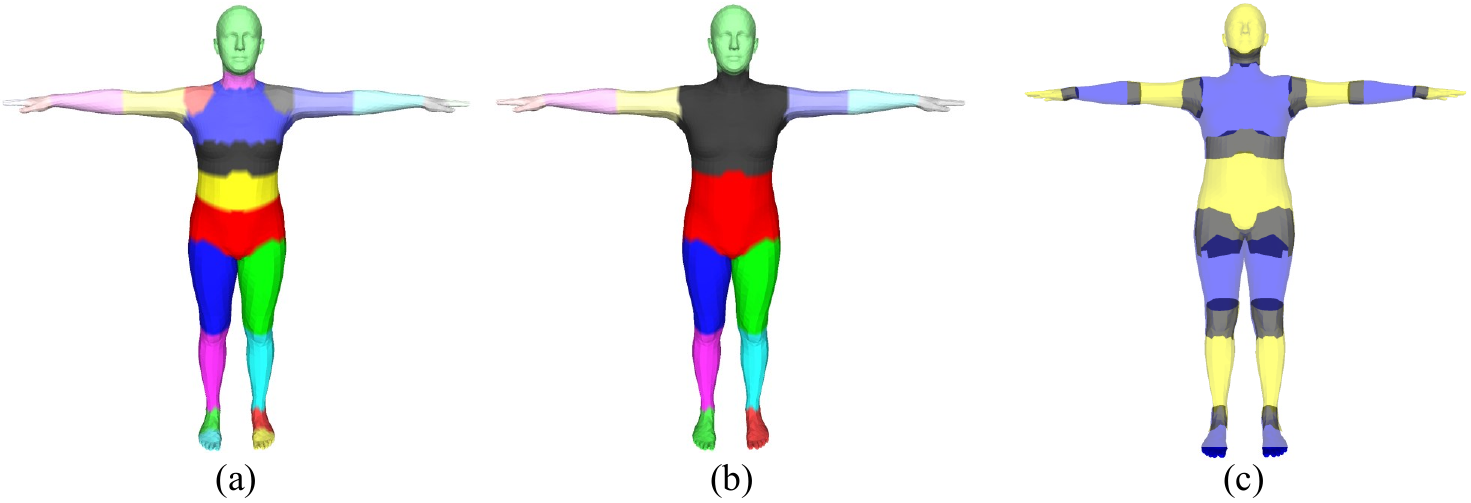}
    \caption{\pca{} template design details.}
    \label{fig:sup1}
\end{figure*}

As shown in \cref{fig:sup1}, we use the SMPL template to design our \pca{} template according to the following three steps. 
First, we initiate with a segmentation of the mesh according to Eq. (1) from the main paper, showcased in \cref{fig:sup1}\textcolor{red}{a}. This segmentation creates 23 distinct segments, corresponding to the number of bones in the SMPL skeleton, to ensure each segment is aligned with a specific bone.
Second, adjacent segments that exhibit nearly-rigid behavior are grouped together to form larger, coherent sections of the mesh, leading to the grouped segmentation shown in \cref{fig:sup1}\textcolor{red}{b}.
Finally, to ensure smooth transitions and coverage around joint areas, we apply graph dilation. This process expands each segment to slightly overlap with its neighbors, enhancing the mesh's flexibility and realism around body joints. For a given body part $p$, its $n$-nearest-neighbor dilated mesh mask is determined by:
\begin{equation}
    m_{pn} = A^nm_p,
\end{equation}
where $m_p \in \{0,1\}^{N}$ is the initial vertex mask for part $p$, indicating whether each of the $N$ vertices in the SMPL template belongs to part $p$ ($m_p=1$ for inclusion, and $m_p=0$ otherwise). The matrix $A$ represents the adjacency matrix of the SMPL mesh graph, and $m_{pn} \in \{0,1\}^{N_p}$ denotes the vertex mask of part $p$ after applying $n$-nearest-neighbor dilation, with $N_p$ being the dilated number of vertices. In this paper, the dilation parameter is empirically set to $n=5$, resulting in the dilated mesh depicted in \cref{fig:sup1}\textcolor{red}{c}. This step ensures each body part mesh extends to overlap with adjacent parts, facilitating a more integrated and natural representation of the human body parts in the final HPPM template.

\begin{table*}[t]
    \centering
    \resizebox{\linewidth}{!}{
    \begin{tabular}{c|c|ccc|ccc|c|ccc|ccc|ccc}
    \hline
        \diagbox{Part Names}{Joints} & Pelvis & \makecell{Right \\ Hip} & \makecell{Right \\ Knee} & \makecell{Right \\ Ankle} & \makecell{Left \\ Hip} & \makecell{Left \\ Knee} & \makecell{Left \\ Ankle} & Torso & Neck & Nose & Head & \makecell{Left \\ Shoulder} & \makecell{Left \\ Elbow} & \makecell{Left \\ Wrist} & \makecell{Right \\ Shoulder} & \makecell{Right \\ Elbow} & \makecell{Right \\ Wrist} \\ \hline \hline
        Abdomen & \checkmark & \checkmark & ~ & ~ & \checkmark & ~ & ~ & \checkmark & ~ & ~ & ~ & ~ & ~ & ~ & ~ & ~ & ~ \\ \hline
        Left Thigh & ~ & ~ & ~ & ~ & \checkmark & \checkmark & ~ & ~ & ~ & ~ & ~ & ~ & ~ & ~ & ~ & ~ & ~ \\ 
        Right Thigh & ~ & \checkmark & \checkmark & ~ & ~ & ~ & ~ & ~ & ~ & ~ & ~ & ~ & ~ & ~ & ~ & ~ & ~ \\  \hline
        Left Calf & ~ & ~ & ~ & ~ & ~ & \checkmark & \checkmark & ~ & ~ & ~ & ~ & ~ & ~ & ~ & ~ & ~ & ~ \\ 
        Right Calf & ~ & ~ & \checkmark & \checkmark & ~ & ~ & ~ & ~ & ~ & ~ & ~ & ~ & ~ & ~ & ~ & ~ & ~ \\  \hline
        Chest & ~ & ~ & ~ & ~ & ~ & ~ & ~ & \checkmark & \checkmark & ~ & ~ & \checkmark & ~ & ~ & \checkmark & ~ & ~ \\  \hline
        Left Foot & ~ & ~ & ~ & \checkmark & ~ & ~ & ~ & ~ & ~ & ~ & ~ & ~ & ~ & ~ & ~ & ~ & ~ \\ 
        Right Foot & ~ & ~ & ~ & ~ & ~ & ~ & \checkmark & ~ & ~ & ~ & ~ & ~ & ~ & ~ & ~ & ~ & ~ \\  \hline
        Head & ~ & ~ & ~ & ~ & ~ & ~ & ~ & ~ & \checkmark & \checkmark & \checkmark & ~ & ~ & ~ & ~ & ~ & ~ \\  \hline
        Left Upper Arm & ~ & ~ & ~ & ~ & ~ & ~ & ~ & ~ & ~ & ~ & ~ & \checkmark & \checkmark & ~ & ~ & ~ & ~ \\
        Right Upper Arm & ~ & ~ & ~ & ~ & ~ & ~ & ~ & ~ & ~ & ~ & ~ & ~ & ~ & ~ & \checkmark & \checkmark & ~ \\  \hline
        Left Forearm & ~ & ~ & ~ & ~ & ~ & ~ & ~ & ~ & ~ & ~ & ~ & ~ & \checkmark & \checkmark & ~ & ~ & ~ \\ 
        Right Forearm & ~ & ~ & ~ & ~ & ~ & ~ & ~ & ~ & ~ & ~ & ~ & ~ & ~ & ~ & ~ & \checkmark & \checkmark \\  \hline
        Left Hand & ~ & ~ & ~ & ~ & ~ & ~ & ~ & ~ & ~ & ~ & ~ & ~ & ~ & \checkmark & ~ & ~ & ~ \\ 
        Right Hand & ~ & ~ & ~ & ~ & ~ & ~ & ~ & ~ & ~ & ~ & ~ & ~ & ~ & ~ & ~ & ~ & \checkmark \\ \hline
    \end{tabular}
    }
    \caption{Part-joint Correspondence Table. }
    \label{tab:sup1}
\end{table*}
\section{\pca{} Part-joint Correspondence}
The correspondence between \pca{} body parts and their associated joints is outlined in \cref{tab:sup1}. We utilize a ``checkmark'' ($\checkmark$) to denote which joints correspond to each \pca{} body part. Specifically, for every \pca{} body part, only those joints marked with a checkmark are included in its joint regressor, establishing a clear linkage between body segments and their respective joints for accurate pose representation.

\section{Evaluation Metrics}
MPVE is defined as:
\begin{equation}
    \textit{MPVE} = \frac{1}{M}\sum(\|\hat{v} - v\|),
\end{equation}
where $M$ is the number of data cases, $\hat{v}$ corresponds to the estimated vertex locations, and $v$ represents the ground-truth vertex locations. MPJPE is defined as:
\begin{equation}
    \textit{MPJPE} = \frac{1}{M}\sum(\|\hat{J} - J\|),
\end{equation}
where $M$ is the number of data cases, $\hat{J}$ is the estimated joint locations, and $J$ is the ground-truth joint locations.

\section{Adaptiveness on Full-body Visible Inputs}

Primarily designed for scenarios with partially visible humans, our method demonstrates notable adaptability to scenarios featuring fully visible body inputs. This capability is showcased without requiring any modifications to the original approach. As indicated in \cref{tab:fullbody}, the performance of our method in full-body visibility images remains competitive with that of previous methods, highlighting the flexibility and efficacy of our approach.

\begin{table*}[t]
    \centering
    \caption{Adaptiveness on full-body visible inputs. As shown in the table, our method can also get reasonable results on full-body visible input.}
    \begin{tabular}{l|cc|c}
    \hline
        \multirow{2}{*}{Model} & \multicolumn{2}{c|}{3DPW~\cite{pw3d}}  & Human3.6M\cite{pw3d} \\
        \cline{2-4}
        & MPJPE/mm$\downarrow$ & MPVE/mm$\downarrow$ & MPJPE/mm$\downarrow$ \\ 
        \hline
        HMR~\cite{kanazawaHMR18} & 130 & - & 88.0 \\
        HMMR~\cite{kanazawa2019learning} & 116.5 & 139.3 & 83.7\\
        BMP~\cite{zhang2021body} & 104.1 & 119.8 & - \\
        VIBE~\cite{kocabas2020vibe} & 93.5 & 113.4 &  65.6\\
        \hline 
        \method{}(Ours) & 100.5 & 123.7 & 64.6 \\ \hline
    \end{tabular}
    \label{tab:fullbody}
\end{table*}

\section{Future Works}
Though our \pca{} can express some deformation on hands and faces, its design could be extended, \eg, by leveraging the SMPL-X body template~\cite{SMPL-X:2019}. This extension would increase the accuracy of hand poses and facial expressions.
Besides, a part-detection or segmentation network such as~\cite{redmon2016you,redmon2018yolov3} could be added to automatically detect which part is visible. The part human reconstruction also has the potential to be used in general articulated objects \cite{zhangs3o, zhang2024magicpose4d}, or to help medical image analysis \cite{xie2024mh1, xie2024mh2, xie2024pflfe, xi2023chain} and action analysis \cite{zhai2023soar, zhu2021enriching, zhu2022learning, zhai2023towards, zhai2020two, xi2023open} as an input condition. Furthermore, we can improve more realistic details by reforming metrics in \cite{luan2024spectrum} to loss using \cite{qi2020simple, qi2021stochastic, qi2022attentional}, or volume-based approaches such as \cite{song2022pref, lou2024darenerf}.

\bibliographystyle{splncs04}
\bibliography{main}
\end{document}